\documentclass[letterpaper, 10 pt, conference]{ieeeconf}  

\IEEEoverridecommandlockouts                              

\usepackage{graphics} 
\usepackage{epsfig} 
\usepackage{mathptmx} 
\usepackage{times} 
\usepackage{amsmath} 
\usepackage{amssymb}  
\usepackage{graphicx}
\usepackage{cite}
\usepackage{url}
\usepackage{multirow}
\title{\LARGE \bf
LNN-Fly: Continuous-Time UAV Navigation for Robust Obstacle Avoidance under Timing Mismatch
}

\author{Yulin Huang, Shaojie Chen, Di Feng, Jiahao Wang,
Ping Liu$^\dagger$, and Jianxiao Zou
\thanks{*This work was supported by the National Key R\&D Program of China (No. 2021YFB2206602), the Shenzhen Science and Technology Program (No. ZDCYKCX20250901092001001, No. SYSRD 20250529114504006, and No. JCYJ20240813114206010), and the Graduate Teaching Research and Reform Project of the Shenzhen Institute for Advanced Study (No. JYJG2026007).}%
\thanks{Yulin Huang, Shaojie Chen, Di Feng, and Jiahao Wang are master's students with the Shenzhen Institute for Advanced Study, University of Electronic Science and Technology of China, Shenzhen, Guangdong 518000, China.
Ping Liu and Jianxiao Zou are with the Shenzhen Institute for Advanced Study, University of Electronic Science and Technology of China, Shenzhen, Guangdong 518000, China.
Corresponding author: Ping Liu ({\tt\small liup81@uestc.edu.cn}).}%
}

\begin{document}

\maketitle
\thispagestyle{empty}
\pagestyle{empty}

\noindent{\small\copyright\ 2026 IEEE. Personal use of this material is permitted. Permission from IEEE must be obtained for all other uses, in any current or future media, including reprinting/republishing this material for advertising or promotional purposes, creating new collective works, for resale or redistribution to servers or lists, or reuse of any copyrighted component of this work in other works.}

\vspace{1em}

\begin{abstract}
End-to-end unmanned aerial vehicle (UAV) navigation can achieve impressive agility in simulation, yet its obstacle-avoidance behavior often degrades after deployment because the policy must tolerate simulator mismatch, sensing irregularity, and variable-rate control. These effects are especially dangerous in cluttered environments, where stale observations or short control irregularities can directly lead to collisions. We present LNN-Fly, a deployment-oriented continuous-time navigation policy for LiDAR-based UAV obstacle avoidance. The policy combines a dynamic-programming-inspired structured recurrent update, explicit conditioning on the elapsed control interval $\Delta t$, and an input-driven adaptive forgetting gate that refreshes stale latent state near hazards while preserving consistency during sustained maneuvers. It is trained with differentiable rollouts that incorporate deployment-relevant sensing and timing perturbations. In simulation, LNN-Fly improves obstacle-avoidance performance in the tested settings and shows better tolerance to reduced control frequency, sparse observations, and control-period jitter. It also transfers zero-shot from a simplified differentiable simulator to a physical quadrotor. In indoor cross-frequency real-world tests, the system achieves 100\% success over 20 flights, while policy inference has a median latency of 0.514\,ms on a desktop graphics processing unit (GPU) and about 2.5\,ms on the onboard central processing unit (CPU), with onboard P95 latency below 30\,ms.
\end{abstract}

\section{INTRODUCTION}

End-to-end learning for unmanned aerial vehicle (UAV) navigation has advanced rapidly, with recent breakthroughs in champion-level high-speed racing\cite{ref25,ref9}, agile flight in forests and the wild\cite{ref5,ref14,ref15}, and fully autonomous indoor navigation for search-and-rescue\cite{ref1,ref4}. These results demonstrate the potential of learned flight policies, but they do not fully resolve a more practical requirement: reliable, collision-free navigation in cluttered environments under real onboard constraints. In many target applications, including industrial inspection, indoor security, search-and-rescue\cite{zhang2024shrinking}, and wildfire monitoring\cite{hasan2022uav}, safety and robustness matter more than pushing the maximum achievable speed. Despite substantial progress, end-to-end policies still degrade noticeably when transferred across simulators, sensing pipelines, and onboard control stacks\cite{ref19}. We focus on two coupled challenges that are especially important for deployment:

\textbf{Challenge 1: Strong and transfer-stable obstacle 
avoidance.}
Existing end-to-end policies often prioritize speed or trajectory completion, while obstacle avoidance is treated as a downstream consequence of good control. In practice, however, dense clutter leaves little margin for error: small discrepancies in physics models, sensor processing, or flight-control pipelines can accumulate into collisions after transfer\cite{wu2025model}. For deployment, the policy must therefore do more than succeed in its training simulator; it must preserve strong obstacle-avoidance behavior when the evaluation stack differs from the training setup.

\begin{figure}[t]
  \centering
  \includegraphics[width=\columnwidth]{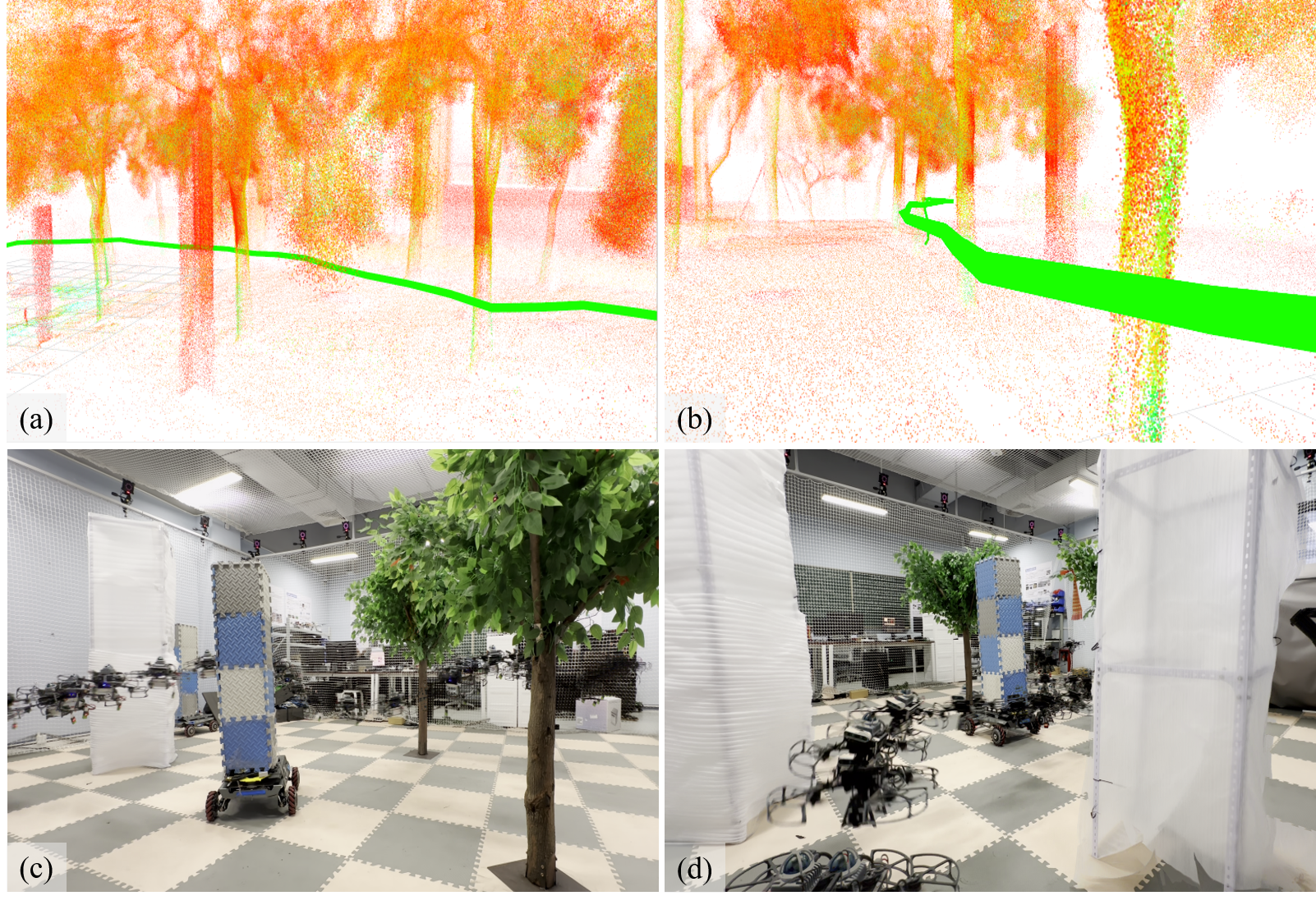}
    \caption{\textbf{Real-world evaluations in diverse environments.} Subfigures (a) and (b) show real-flight experiments in an outdoor forest environment, while (c) and (d) show real-flight experiments in a narrow indoor laboratory corridor. The results demonstrate robust navigation performance across highly cluttered natural terrain and confined man-made spaces.}
    
  \label{fig:realfly}
\end{figure}

\textbf{Challenge 2: Robustness to timing mismatch in real 
control loops.}
Real onboard systems rarely maintain a fixed control period: irregular sensor sampling, rate drops, and scheduling jitter are common on resource-constrained edge computers. Yet many learned sequence policies are trained and deployed as if control updates arrived at a constant rate. When the effective $\Delta t$ changes at deployment, latent-state evolution and physical motion can become misaligned, producing unsafe commands\cite{ref20,ref35,rubanova2019latent}. This mismatch is particularly dangerous during close-proximity avoidance, where even brief control irregularities can directly lead to collisions.

To address these challenges jointly, we propose \textbf{LNN-Fly}, a lightweight continuous-time framework for UAV obstacle avoidance under deployment mismatch. It combines a recurrent update explicitly conditioned on the elapsed control interval $\Delta t$ with an input-driven adaptive forgetting mechanism that refreshes stale latent state near hazards while preserving consistency during sustained maneuvers.

Architecture alone is not sufficient for robust deployment. We therefore train the policy with differentiable rollouts that expose it to sensing and timing perturbations relevant to real operation, including asynchronous observations, control-period variation, and actuation delay\cite{ref16}. This yields a compact policy whose design, training setting, and evaluation protocol are all organized around the same deployment bottleneck: maintaining reliable obstacle avoidance when timing and system conditions differ from those seen during nominal training.

This work treats continuous-time recurrence as a task-structured component for UAV obstacle avoidance: the recurrent state should evolve with time while remaining tied to local obstacle context. The dynamic-programming analogy motivates recursive local consistency rather than a Bellman optimization claim, and is combined with measured-time conditioning, input-driven memory refresh, and deployment-oriented evaluation.

In summary, our main contributions are as follows:
\begin{itemize}
     \item \textbf{A timing-aware continuous-time policy:} We propose LNN-Fly, a lightweight recurrent policy for UAV obstacle avoidance with explicit $\Delta t$ conditioning and input-driven adaptive forgetting.
    \item \textbf{A deployment-oriented training and evaluation pipeline:} We train with differentiable rollouts under sensing and timing perturbations, and evaluate across simulator shifts, temporal mismatch, and physical-quadrotor deployment.
    \item \textbf{Zero-shot transfer to cluttered real-world flight:} We integrate the policy into a LiDAR-based navigation system and validate it in Gazebo+PX4 and physical flight, including outdoor forest and narrow indoor corridor tests (Fig.~\ref{fig:realfly}).
\end{itemize}

\section{RELATED WORK}
\label{sec:related_work}

\subsection{End-to-End UAV Navigation and Obstacle Avoidance}

Modular planning-and-control pipelines construct explicit environmental representations and generate trajectories via optimization\cite{zhou2019robust,zhou2020ego,zhou2021ego}, achieving reliable flight but introducing cumulative latency and error propagation that limit responsiveness in dense clutter.

End-to-end learning sidesteps these limitations by mapping sensor observations directly to control commands, spanning indoor autonomy, search-and-rescue, trail following, and self-supervised avoidance\cite{ref1,ref2,ref4}, as well as high-speed agile flight\cite{ref5,ref9,ref12,ref13,ref15,ref25,ref32,romero2025dream,ref33}. 

Despite this progress, most of these methods primarily optimize for speed or trajectory tracking, while strong obstacle-avoidance capability and its stability under deployment shifts receive comparatively less attention. In practice, avoidance performance often degrades significantly when policies are transferred across simulators, sensors, or flight stacks\cite{ref19,wu2025model}, and learning-based control can further suffer from limited interpretability and hard-to-diagnose failure modes under distribution shifts\cite{ref18}. Existing remedies include domain randomization, scene transfer, adaptation, improved dynamics modeling, and explicit safety mechanisms\cite{ref23,ref11,ref26,ref27,finn2017model,kumar2021rma,ref28,ref29,ref30}. These directions are useful, but transfer-stable avoidance remains difficult when sensing and control timing also vary. We therefore treat obstacle avoidance as the primary objective and evaluate transfer behavior under deployment-relevant shifts.

\subsection{Timing Mismatch and Continuous-Time Policies}

Timing variability (e.g., irregular sampling, rate drops, and scheduling jitter) is pervasive on resource-constrained onboard computers. Such irregular updates challenge discrete sequence learners. Traditional discrete-time sequence models\cite{ref20,ref21} implicitly assume a constant timestep, causing performance to degrade when the effective $\Delta t$ fluctuates at deployment. Continuous-time models, such as Neural ODEs\cite{ref36}, Neural CDEs\cite{kidger2020neural}, and liquid/closed-form continuous-time networks\cite{ref38,ref37}, address this by providing principled mechanisms to evolve hidden states under variable $\Delta t$. 

Recent studies have leveraged liquid networks for robust flight navigation and sim-to-real transfer\cite{ref34}. These models provide useful mechanisms for irregular timing, but they are often used as generic temporal filters. Our focus is narrower and task-structured: we shape the recurrent dynamics around obstacle-avoidance behavior, where stale perception, local target consistency, and variable control intervals jointly affect safety. LNN-Fly combines a dynamic-programming-inspired consistency view with a $\Delta t$-conditioned update and input-driven forgetting to improve tolerance across the tested simulator, sensing, and timing shifts.

\section{METHOD}
\begin{figure*}[t]
  \vspace{2mm}
  \centering
  \includegraphics[width=0.84\textwidth]{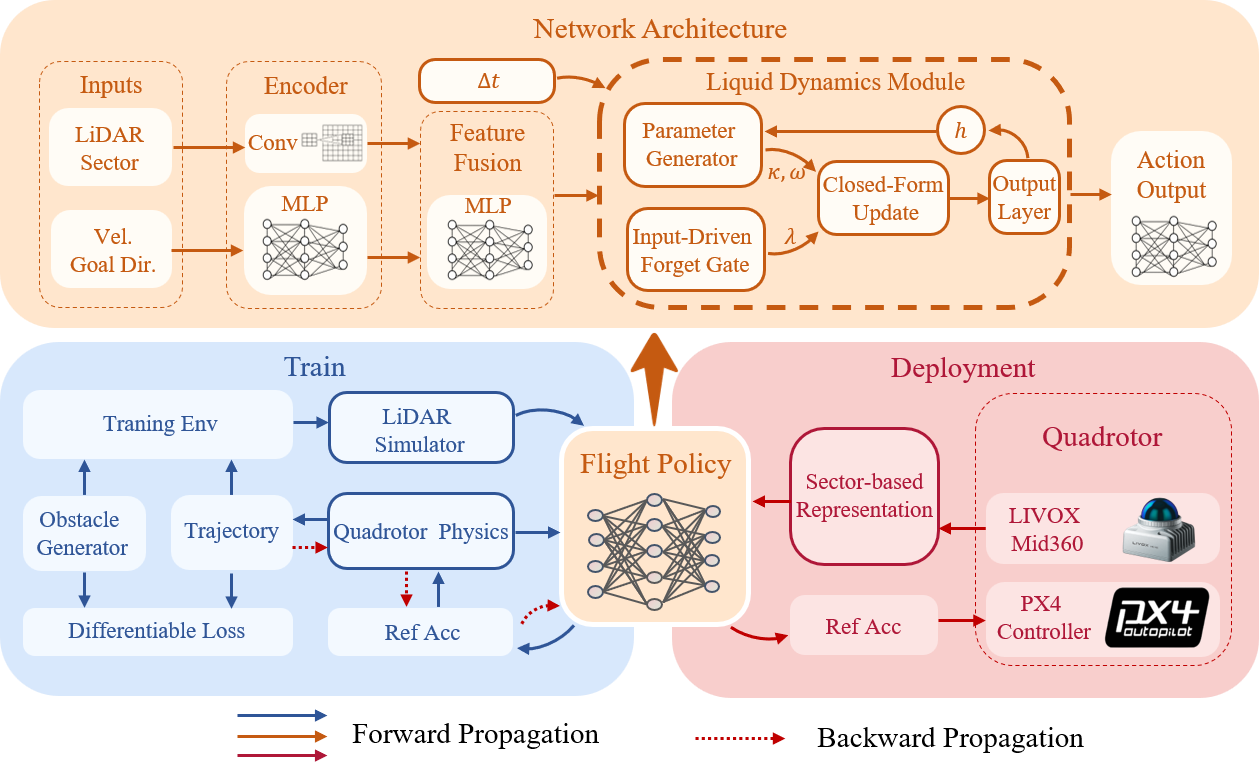}
\caption{\textbf{Overview of the LNN-Fly framework.} The system processes LiDAR sectors\cite{ref40} and state cues via a lightweight encoder and a Closed-form Continuous-time (CfC) core. An input-driven forgetting mechanism dynamically modulates the temporal scale to balance smoothness and agility, optimized end-to-end via differentiable physics.}
\label{fig:framework}
\end{figure*}

\subsection{Problem Formulation and System Overview}

We formulate the UAV navigation task as a sampled-data Partially Observable Markov Decision Process (POMDP) under irregular sampling intervals and potential deployment shifts. 
Let $s_k \in \mathcal{S}$ denote the physical state (position, velocity, orientation), $y_k \in \mathcal{O}$ the observation (LiDAR scans, inertial measurement unit (IMU) data), and $u_k \in \mathcal{U}$ the control action. The system evolves as $s_{k+1} = \Phi_d(s_k, u_k, \Delta t_k)$ with $y_k = \Psi_d(s_k)$, where $d$ denotes the operating domain and $\Delta t_k = t_{k+1} - t_k$.

In real deployment, both the dynamics and sensing models, as well as timing, can deviate from simulation ($\Phi_{real} \neq \Phi_{sim}$, $\Psi_{real} \neq \Psi_{sim}$, and variable $\Delta t_k$). Our goal is to learn a continuous-time policy $\pi_\theta(h_k, y_k, \Delta t_k)$ that maps a recurrent belief surrogate $h_k$ to actions $u_k$ while remaining robust to these discrepancies.

To this end, we propose the \textbf{LNN-Fly Navigation Framework} (Fig.~\ref{fig:framework}) with three components: a compact continuous-time policy architecture (Sec.~III-B and III-C), an end-to-end differentiable-physics learning loop (Sec.~III-D), and a real-world closed-loop deployment system enabling zero-shot transfer.

\textbf{Policy Network Architecture.}
To meet lightweight onboard execution requirements, we adopt a compact encoder--recurrent--head design.
The LiDAR input ($60 \times 15$ angular resolution) is reshaped to $[B,1,15,60]$ and encoded by a 4-layer convolutional neural network (CNN) ($1 \rightarrow 12 \rightarrow 24 \rightarrow 36$) followed by a 2-layer multilayer perceptron (MLP), producing a 64-d feature.
The 10-d state vector is encoded by an MLP ($10 \rightarrow 48 \rightarrow 48$).
These features are concatenated into a 112-d fused vector and fed to the Closed-form Continuous-time (CfC) temporal unit (hidden size 32).
Finally, a policy head ($32 \rightarrow 32 \rightarrow 6$) outputs a 6-D command $u_k=[a_k^\top,\,c_k^\top]^\top$, where $a_k \in \mathbb{R}^3$ is the desired body/world-frame acceleration and $c_k \in \mathbb{R}^3$ is a velocity-feedback correction term used to compensate velocity drift in the absence of odometry.

\subsection{Dynamic-Programming-Inspired Neural Liquid Dynamics}

Standard recurrent neural networks (RNNs) (e.g., long short-term memory networks (LSTMs)) are typically applied with discrete updates, making them sensitive to irregular timing and stale latent state during deployment. Rather than using a generic recurrent transition, we parameterize the hidden-state dynamics with an explicit contraction form that is compatible with variable control intervals.

\textit{Formulation via Structured Continuous Consistency:}
Inspired by recursive planning, we interpret the hidden state $h(t)$ as a compact latent summary that should remain consistent with an observation-conditioned local target. Let $\kappa(x_t,h_t)$ denote this learned local target, predicted from the current observation and latent context. The dynamic-programming analogy is used only as a design motivation for this contraction structure; the model does not optimize a Bellman objective or perform value iteration. Instead of adding an explicit consistency supervision term, we encode this preference directly in the recurrent dynamics. Define the local consistency energy
\begin{equation}
\mathcal{E}(t)=\frac{1}{2}\|h(t)-\kappa(x_t,h_t)\|_2^2.
\label{eq:consistency_energy}
\end{equation}
Ignoring higher-order dependence of $\kappa$ on $h$ within an infinitesimal interval, descent dynamics on $\mathcal{E}$ with an additional input-driven leakage term yield
\begin{equation}
\frac{d h}{dt}
=
\underbrace{\omega(x,h)\odot\big(\kappa(x,h)-h\big)}_{\text{Structured Target Contraction}}
-
\underbrace{\lambda(x)\odot h}_{\text{Stimulus-Driven Discounting}}.
\label{eq:ode_dynamics}
\end{equation}
This parameterization restricts each hidden channel to a contraction form toward a local target, rather than an unconstrained recurrent rewrite.

\textit{Closed-Form Discrete Transition:}
To handle irregular $\Delta t_k$, we solve (\ref{eq:ode_dynamics}) under a piecewise-constant approximation on $[t_k,t_{k+1})$:
\begin{equation}
\kappa(t)\approx \kappa_k,\quad \omega(t)\approx \omega_k,\quad \lambda(t)\approx \lambda_k,
\label{eq:piecewise_const}
\end{equation}
which gives the linear ODE
\begin{equation}
\dot h(t)=-(\omega_k+\lambda_k)\odot h(t)+\omega_k\odot \kappa_k.
\label{eq:linear_ode}
\end{equation}
We explicitly inject the time interval by conditioning the network parameters on $\delta_k = \log(\Delta t_k + \epsilon)$.
The discrete state update is:
\begin{align}
h_{k+1} &= \alpha_k \odot h_k + (1-\alpha_k) \odot \frac{\omega_k}{\omega_k + \lambda_k + \epsilon_s} \odot \kappa_k, \\
\alpha_k &= \exp\left( -(\omega_k + \lambda_k) \Delta t_k \right).
\end{align}
Here, $\kappa_k,\omega_k,\lambda_k,\alpha_k,h_k \in \mathbb{R}^{32}$ are channel-wise hidden-state quantities, $\odot$ denotes the Hadamard product, $\epsilon_s$ is a small numerical constant, and $\alpha_k$ acts as a time-aware gating factor.
The bounds $\lambda_{floor}$ and $\lambda_{max}$ are not separately optimized; they serve as conservative lower/upper limits for adaptive forgetting. We use $\lambda_{floor}=0$ to preserve near-memory retention in benign conditions and a finite $\lambda_{max}$ to avoid unstable over-forgetting; empirically, performance is stable under moderate changes of these bounds, except when $\lambda_{max}$ is set too small.
In our implementation, $\delta_k$ scales the generation of $\kappa_k$, $\omega_k$, and the forgetting gate $\lambda_k$, as detailed in Sec.~III-C. The effective neural time constant $\tau_k = (\omega_k + \lambda_k)^{-1}$ therefore adapts jointly to the input features and the physical time elapsed, allowing the recurrent update to remain synchronized with variable-rate control.

\textit{How the Inductive Bias Enters Learning:}
During training, we do \emph{not} directly minimize a supervised consistency residual. Instead, Backpropagation Through Time (BPTT) updates parameters through the constrained transition:
\begin{equation}
\frac{\partial \mathcal{L}_{total}}{\partial \theta}
=
\sum_k
\frac{\partial \mathcal{L}_{total}}{\partial h_{k+1}}
\frac{\partial h_{k+1}}{\partial (\kappa_k,\omega_k,\lambda_k)}
\frac{\partial (\kappa_k,\omega_k,\lambda_k)}{\partial \theta}.
\label{eq:bptt_chain}
\end{equation}
Moreover, the hidden-to-hidden Jacobian admits the first-order decomposition
\begin{equation}
\frac{\partial h_{k+1}}{\partial h_k}
\approx
\mathrm{Diag}(\alpha_k)
+
\frac{\partial h_{k+1}}{\partial \kappa_k}\frac{\partial \kappa_k}{\partial h_k}
+
\frac{\partial h_{k+1}}{\partial \omega_k}\frac{\partial \omega_k}{\partial h_k},
\label{eq:jacobian_decomp}
\end{equation}
where the dominant self-transition term is $\mathrm{Diag}(\alpha_k)$ with $\alpha_k=\exp(-(\omega_k+\lambda_k)\Delta t_k)\in(0,1]$. Compared with an unconstrained recurrent Jacobian, this transition is centered on an explicit temporal contraction factor, while the remaining terms act as target-shaping corrections. In this sense, the proposed update introduces an implicit consistency bias during BPTT through the contraction-to-$\kappa_k$ transition in (\ref{eq:linear_ode})--(\ref{eq:bptt_chain}).

\begin{figure}[t]
  \centering
  \includegraphics[width=0.95\columnwidth]{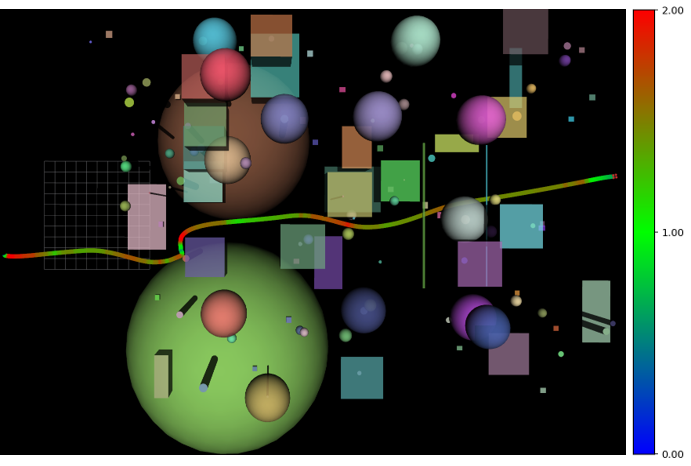}
  \caption{\textbf{Visualization of the adaptive forgetting rate $\lambda_k$.} The trajectory color transitions from blue (low $\lambda_k$) to red (high $\lambda_k$). During sustained avoidance of a large obstacle, $\lambda_k$ remains low to preserve memory and maintain a consistent avoidance direction. After the obstacle is cleared, $\lambda_k$ spikes to rapidly refresh the latent state for the subsequent flight segment.}
  \label{fig:lambda_vis}
\end{figure}

\subsection{Input-Driven Adaptive Time-Scale Regulation}

Instead of using a hand-crafted scheduling rule, we let the current observation directly regulate the forgetting rate. We define $\lambda_k$ as a channel-wise gate driven by fused sensory input $x_k$:
\begin{align}
\lambda_{raw} &= W_{\lambda} x_k + b_{\lambda} + \delta_k \cdot \theta_{\lambda}, \\
\lambda_k &= \mathrm{clip}\!\left(\mathrm{softplus}(\lambda_{raw}) + \lambda_{floor},\ 0,\ \lambda_{\max}\right),
\label{eq:input_mod}
\end{align}
where $W_{\lambda}$, $b_{\lambda}$, and $\theta_{\lambda}$ are learnable parameters. During differentiable-physics training, the gate is optimized only through long-horizon task loss, so its behavior emerges from the need to trade off memory retention and rapid correction. In open regions or during sustained avoidance around the same obstacle, the learned optimum is typically a small $\lambda_k$, which preserves latent momentum and avoids oscillatory overreaction. Near abrupt hazards or under deployment perturbations, acute proximity cues increase $\lambda_k$, shorten the effective time constant, and suppress stale hidden-state history.

This mechanism supports the behavior studied in the experiments. First, it improves agility without sacrificing smoothness: the policy can remain inertial when the scene is stable and become sharply reactive only when new evidence demands it. Second, it reduces sensitivity to stale latent history under timing or dynamics mismatch by allowing the controller to refresh internal state more aggressively when needed. Fig.~\ref{fig:lambda_vis} visualizes this pattern, with low $\lambda_k$ during sustained avoidance and a post-clearance spike when the latent state is refreshed for the next flight segment.

\begin{figure*}[t]
  \vspace{2mm}
  \centering
  \includegraphics[width=0.93\textwidth]{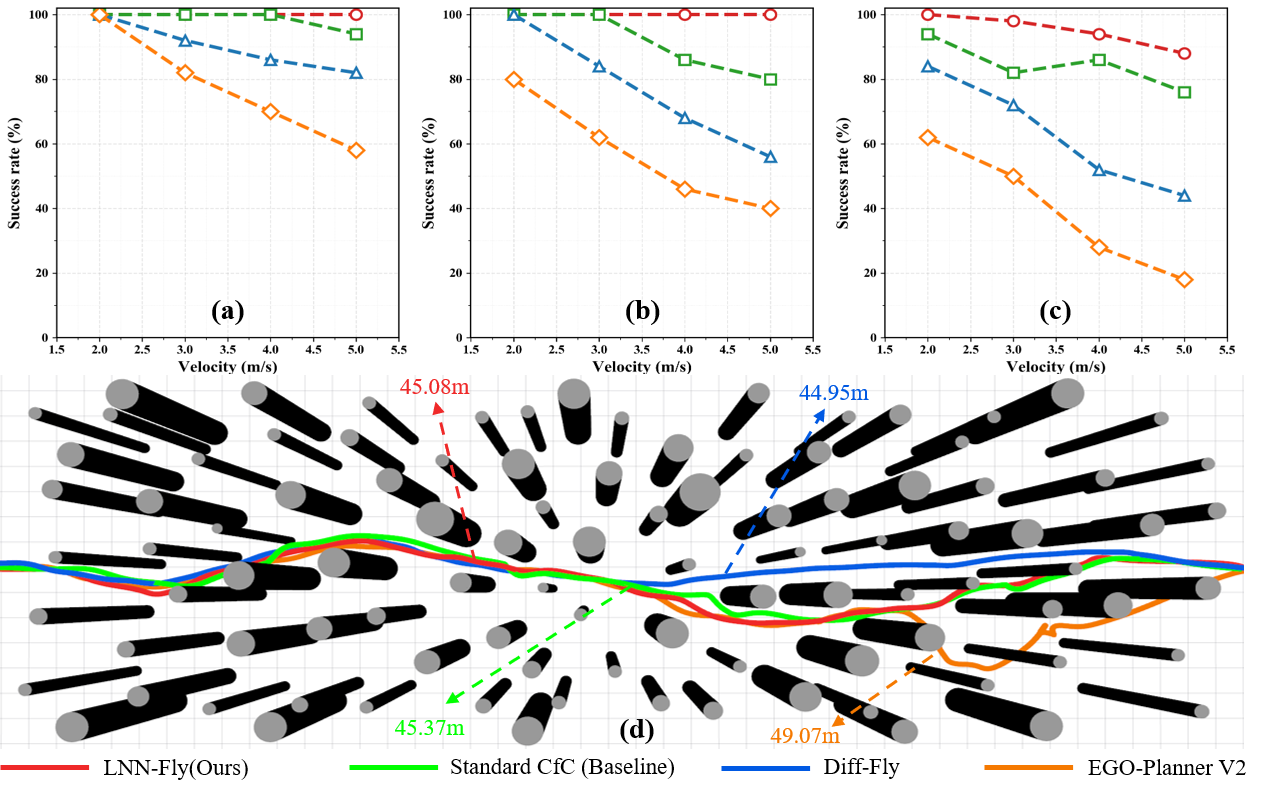}
  \caption{\textbf{Benchmark results with previous systems under different maximum speeds.}
Figures (a)--(c) illustrate the success-rate comparisons in Scenarios I, II, and III, respectively (50 trials per speed). 
(d) Example trajectory visualizations in Scenario II with the speed constraint of $4.0$\,m/s, where the flight distances (trajectory lengths) are provided.}

\label{fig:benchmark}
\end{figure*}

\subsection{Learning via Differentiable Physics}

We train the compact policy with differentiable physics, which keeps optimization efficient while exposing the recurrent dynamics to deployment-relevant timing and sensing effects.

\textit{Forward Propagation:}
The policy outputs an acceleration command $\tilde{a}_k$ and a correction term $\tilde{c}_k$, which are combined with gravity compensation to form the final control $u_k$. We then roll out the differentiable dynamics $s_{k+1} = \Phi(s_k, u_k, \Delta t_k)$ using the same sampled control interval $\Delta t_k$ as the recurrent update, so latent-state evolution and physical propagation remain temporally consistent during training. To narrow sim-to-real mismatch, the rollout preserves simulation-real shared invariants, including sectorized LiDAR geometry, local obstacle layout, and control semantics, while injecting deployment-relevant disturbances such as LiDAR noise and drift, partial observations, asynchronous sensing, control-period jitter, and actuation latency. The goal is not to reproduce the full real world, but to emphasize features that are shared across simulation and hardware while reducing sensitivity to deployment-specific perturbations.

\textit{Objective Function:}
We unroll the system for a horizon $T$ and optimize a composite loss $\mathcal{L}_{total}$, defined in (\ref{eq:total_loss}), via Backpropagation Through Time (BPTT):
\begin{equation}
\mathcal{L}_{total} = w_1 \mathcal{L}_{track} + w_2 \mathcal{L}_{safe} + w_3 \mathcal{L}_{smooth} + w_4 \mathcal{L}_{reg}.
\label{eq:total_loss}
\end{equation}
where $w_i$ are scalar weights.
\begin{itemize}
    \item $\mathcal{L}_{track}$: Penalizes deviations from target velocity, heading direction, and altitude.
    \item $\mathcal{L}_{safe}$: Enforces collision avoidance using a smooth barrier function $\phi(d) = \text{softplus}(d_{\text{safe}} - d)$ on the distance to nearest obstacles, weighted by closing speed, where $d_{\text{safe}}$ is the safety threshold.
    \item $\mathcal{L}_{smooth}$: Penalizes control effort (acceleration) and higher-order derivatives (jerk, snap) to ensure feasible trajectories.
    \item $\mathcal{L}_{reg}$: Regularizes the input-driven forgetting distribution (via L1, mean target, and soft-range penalties) to maintain stable long-term memory while allowing agile reactions.
\end{itemize}
These objectives supervise the policy only through the differentiable rollout and the structured recurrent transition of Sec.~III-B, shaping $\kappa$, $\omega$, and $\lambda$ so that the induced latent dynamics improve long-horizon tracking, safety, and smoothness.

\section{EXPERIMENTS}

\subsection{Setup and Implementation}
\label{sec:exp_setup}

To assess scenario generalization, training is performed in a differentiable simulator with a simplified point-mass quadrotor model, and evaluation is performed in a Gazebo+PX4 software-in-the-loop (SITL) stack, creating shifts in simulator implementation, sensing, and dynamics. Training and simulation-based tests run on a workstation with an Intel i7-13700 central processing unit (CPU) and an NVIDIA RTX 5080 graphics processing unit (GPU), while real flights use an onboard Intel Core Ultra 7 155H CPU only. We optimize with AdamW (learning rate $5\times 10^{-4}$) for 10,000 iterations using rollout horizon $T=150$ and batch size 128. Training includes a 2-step action buffer, perturbed control intervals around 30\,Hz, and asynchronous LiDAR updates to expose the policy to latency and timing jitter.

\subsection{Benchmark with Previous Systems}
\label{sec:benchmark}

\textbf{Evaluation setup and baselines.}
Evaluations run in a $35{\times}10{\times}6$\,m Gazebo+PX4 SITL workspace (quadrotor safety envelope ${\approx}0.335$\,m) across three static scenarios with increasing obstacle density and occlusion. Scenario I contains 100 obstacles with radius $[0.06, 0.50]$\,m ($7.17\%$ coverage, $1.09{\pm}0.36$\,m spacing), Scenario II contains 100 obstacles with radius $[0.16, 0.60]$\,m ($9.91\%$ coverage, $0.98{\pm}0.38$\,m spacing), and Scenario III contains 130 obstacles with radius $[0.06, 0.55]$\,m ($9.39\%$ coverage, $0.88{\pm}0.38$\,m critical spacing). We compare against \textbf{Standard CfC}\cite{ref37}, \textbf{Diff-Fly}\cite{ref14}, and \textbf{EGO-Planner V2}\cite{zhou2021ego}. Specifically, \textbf{Standard CfC} is a reconstructed baseline adapted from \cite{ref37} with a parameter count comparable to our method for a fair comparison. Unless otherwise noted, \textbf{Standard CfC} and \textbf{Diff-Fly} use the same observations, differentiable-physics training pipeline, rollout horizon, optimization setting, and update frequency as \textbf{LNN-Fly}. \textbf{EGO-Planner V2} is provided with the full observations used in its original pipeline and is evaluated at the same update frequency.

\textbf{Safety, efficiency and domain transfer.}
Across 50 trials per speed in Fig.~\ref{fig:benchmark}(a)--(c), \textbf{LNN-Fly} achieves the best overall combination of success rate and path efficiency as speed and scene density increase. The gap is most visible in Scenario III at $5.0$\,m/s, where \textbf{EGO-Planner V2} drops to $18\%$ success. Relative to the training ablation in Sec.~\ref{sec:ablation_train}, \textbf{Diff-Fly} also shows a larger transfer drop, while \textbf{LNN-Fly} preserves performance more consistently, suggesting that its structured continuous-time design captures obstacle-avoidance cues that remain informative across simulator and flight-stack changes.

\begin{table*}[t]
\vspace{2mm}
\caption{\textbf{Main results under varying control frequencies and perception latencies.} Performance is measured by minimum distance $d_{\min}$ (m, higher is better), distance-below-threshold ratio $d_{\text{below}}$ (lower is better), and overall success rate across 20 trials. \textbf{Hyphens (-)} denote catastrophic failures.}
\label{tab:table1_main}
\centering
\small
\setlength{\tabcolsep}{4pt}
\renewcommand{\arraystretch}{1.05} 
\begin{tabular}{l l ccc c ccc c ccc}
\hline
\multirow{2}{*}{\textbf{Freq.}} & \multirow{2}{*}{\textbf{Model}} & \multicolumn{3}{c}{\textbf{N=1}} & & \multicolumn{3}{c}{\textbf{N=3}} & & \multicolumn{3}{c}{\textbf{N=5}} \\
\cline{3-5} \cline{7-9} \cline{11-13}
& & $d_{\min}$ & $d_{\text{below}}$ & Succ. & & $d_{\min}$ & $d_{\text{below}}$ & Succ. & & $d_{\min}$ & $d_{\text{below}}$ & Succ. \\
\hline
\multirow{4}{*}{30 Hz}
& LNN-Fly                    & 0.368 & 0.068 & 20 && 0.351 & 0.130 & 20 && 0.322 & 0.134 & 18 \\
& LNN-Fly (Fixed $\Delta t$) & 0.343 & 0.073 & 20 && 0.313 & 0.155 & 18 && 0.323 & 0.140 & 19 \\
& Standard CfC               & 0.366 & 0.107 & 20 && 0.359 & 0.123 & 19 && 0.401 & 0.159 & 17 \\
& LSTM                       & 0.389 & 0.156 & 20 && 0.247 & 0.185 & 14 && 0.108 & 0.178 & 11 \\
\hline
\multirow{4}{*}{10 Hz}
& LNN-Fly                    & 0.306 & 0.118 & 20 && 0.325 & 0.159 & 20 && 0.319 & 0.182 & 17 \\
& LNN-Fly (Fixed $\Delta t$) & 0.298 & 0.177 & 20 && 0.276 & 0.161 & 17 && 0.272 & 0.186 & 17 \\
& Standard CfC               & 0.385 & 0.155 & 20 && 0.307 & 0.192 & 17 && 0.358 & 0.158 & 15 \\
& LSTM                       & 0.204 & 0.255 & 20 && 0.156 & 0.193 & 11 && 0.102 & 0.283 & 9  \\
\hline
\multirow{4}{*}{8 Hz}
& LNN-Fly                    & 0.385 & 0.135 & 18 && 0.313 & 0.118 & 14 && 0.277 & 0.196 & 13 \\
& LNN-Fly (Fixed $\Delta t$) & 0.213 & 0.291 & 15 && 0.178 & 0.329 & 11 && 0.230 & 0.365 & 9  \\
& Standard CfC               & 0.337 & 0.108 & 18 && 0.273 & 0.141 & 15 && 0.330 & 0.205 & 13 \\
& LSTM                       & 0.172 & 0.274 & 15 && 0.149 & 0.314 & 8  && 0.147 & 0.373 & 5  \\
\hline
\multirow{4}{*}{5 Hz}
& LNN-Fly                    & 0.281 & 0.116 & 12 && 0.309 & 0.135 & 9  && - & - & - \\
& LNN-Fly (Fixed $\Delta t$) & 0.271 & 0.461 & 5  && - & - & - && - & - & - \\
& Standard CfC               & 0.274 & 0.102 & 10 && 0.262 & 0.147 & 7  && - & - & - \\
& LSTM                       & - & - & - && - & - & - && - & - & - \\
\hline
\end{tabular}
\end{table*}

\begin{table}[t]
\caption{\textbf{Model robustness under control-frequency jitter.} Perception is fixed at $N=3$. Jitter is modeled as a 50\% duty cycle alternating between 30\,Hz and $f_{\text{low}}$ every 2.0\,s.}
\label{tab:freq_jitter}
\centering
\footnotesize 
\setlength{\tabcolsep}{3pt} 
\renewcommand{\arraystretch}{1.1}
\begin{tabular}{l l c c c}
\hline
\textbf{Jitter Mode} & \textbf{Model} & \textbf{$d_{\min}$ (m)} $\uparrow$ & \textbf{$d_{\text{below}}$} $\downarrow$ & \textbf{Success} $\uparrow$ \\
\hline
\multirow{4}{*}{\shortstack{30$\leftrightarrow$8\,Hz\\(Moderate)}}
& LNN-Fly            & \textbf{0.331} & 0.111          & \textbf{17}/20 \\
& LNN-Fly (Fixed $\Delta t$) & 0.271          & 0.156          & 12/20 \\
& Standard CfC               & 0.309          & \textbf{0.109} & 16/20 \\
& LSTM                       & 0.267          & 0.126          & 9/20  \\
\hline
\multirow{4}{*}{\shortstack{30$\leftrightarrow$5\,Hz\\(Severe)}}
& LNN-Fly           & \textbf{0.326} & 0.145          & \textbf{14}/20 \\
& LNN-Fly (Fixed $\Delta t$) & 0.232          & 0.179          & 7/20  \\
& Standard CfC               & 0.295          & \textbf{0.136} & 14/20 \\
& LSTM                       & -              & -              & -  \\
\hline
\end{tabular}
\end{table}

\subsection{Robustness to Temporal Semantic Mismatch} \label{sec:temporal_robustness}

\textbf{Evaluation Protocol.}
We evaluate robustness to temporal mismatch by deploying policies trained at a fixed 30\,Hz in Scenario II at 4.0\,m/s under two timing perturbations. \textbf{Reduced Control Frequency with Sparse Observations:} inference runs at $f \in \{30, 10, 8, 5\}$\,Hz, while the LiDAR observation is refreshed every $N \in \{1, 3, 5\}$ inference steps. For $N>1$, the controller must act on stale observations, emulating perception latency or reduced sensor refresh rates. \textbf{Control-Frequency Jitter:} the control frequency alternates between 30\,Hz and a lower frequency $f_{\text{low}}$, simulating variable compute availability.
To isolate the role of explicit time-awareness, we include \textbf{LNN-Fly (Fixed $\Delta t$)}, which ignores the true $\Delta t_k$ and always integrates with a constant 30\,Hz step ($0.033$\,s).

\textbf{Cross-Frequency Generalization.}
As the control frequency departs from the 30\,Hz training setting, discrete-time baselines degrade sharply (Table~\ref{tab:table1_main}).
At \textbf{5\,Hz} ($\Delta t=0.2$\,s, $6\times$ the training interval), the \textbf{LSTM} exhibits catastrophic failure across all $N$.
Comparing the two continuous-time variants highlights the impact of correct time-conditioning: at 5\,Hz ($N=1$), \textbf{LNN-Fly} remains controllable with 12/20 successes, whereas \textbf{LNN-Fly (Fixed $\Delta t$)} drops to 5/20. This suggests that a continuous-time (ODE) formulation alone is insufficient; conditioning on the true elapsed time $\Delta t$ is necessary to propagate latent state over large inter-step displacements.

\textbf{Robustness to Perception Latency.}
Observation sparsity further stresses temporal alignment. At 30\,Hz, increasing $N$ to 5 reduces the LSTM success rate from 20 to 11, while \textbf{LNN-Fly} remains robust (18/20).
Even at 8\,Hz with severe sparsity ($N=5$), \textbf{LNN-Fly} achieves the highest success rate (13/20), outperforming both \textbf{LNN-Fly (Fixed $\Delta t$)} (9/20) and LSTM (5/20).

\textbf{Robustness to Frequency Jitter.}
Table~\ref{tab:freq_jitter} evaluates stability under time-varying control intervals.
Under severe jitter (30\,Hz $\leftrightarrow$ 5\,Hz), \textbf{LNN-Fly} attains 14/20 successes, doubling \textbf{LNN-Fly (Fixed $\Delta t$)} (7/20).
Because the Fixed $\Delta t$ model integrates with 0.033\,s even when the true interval is 0.2\,s, its internal state quickly desynchronizes from the underlying physics. In contrast, \textbf{LNN-Fly} leverages the varying $\Delta t$ input to adapt its recurrent dynamics online, supporting reliable deployment under unstable timing.



\subsection{Ablation Study on Training Dynamics}
\label{sec:ablation_train}

\begin{figure}[t]
  \centering
  \includegraphics[width=0.98\columnwidth]{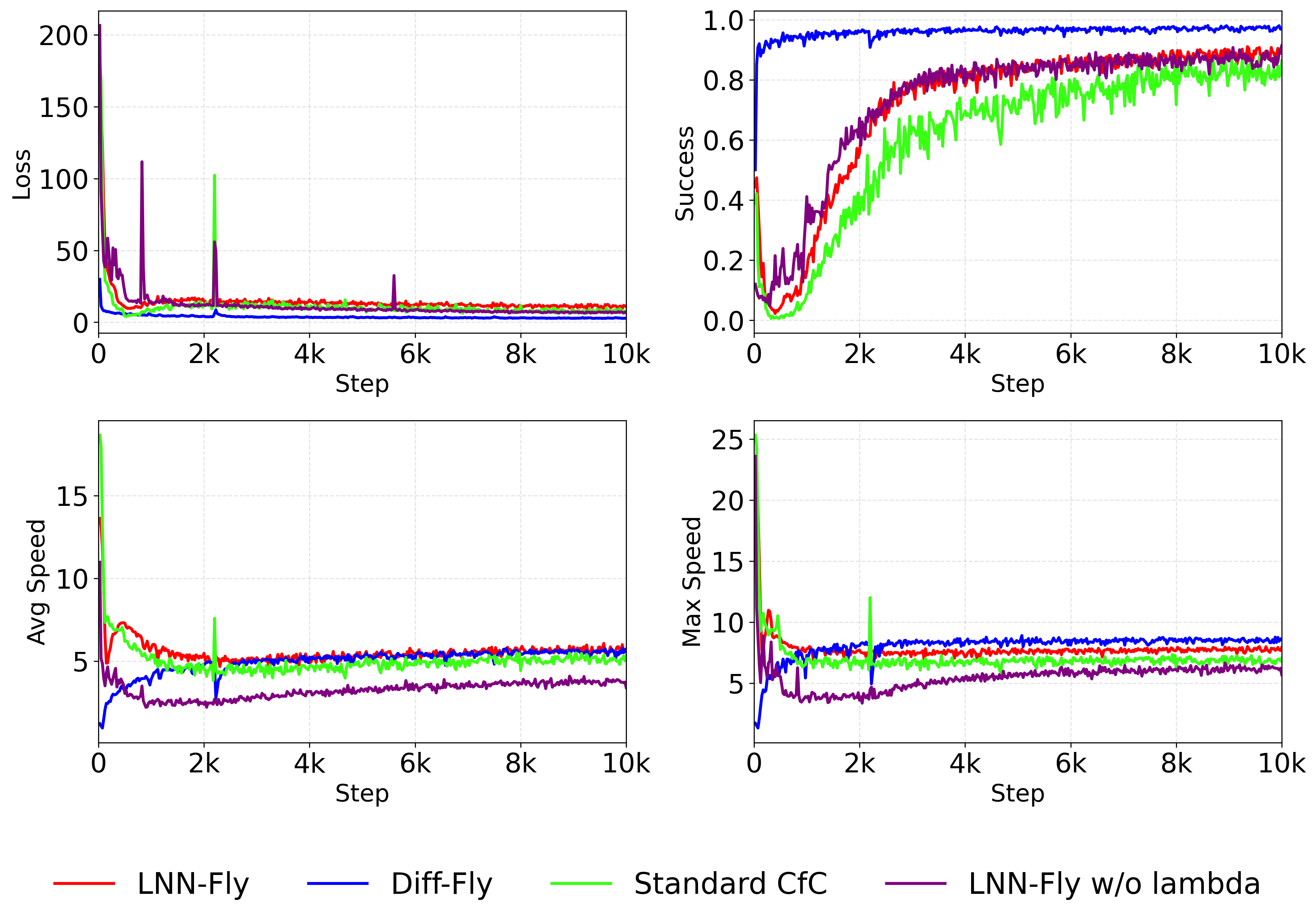}
\caption{\textbf{Comparison of training dynamics over 10,000 steps.} From left to right and top to bottom, the subplots show loss, success rate, maximum speed, and average speed, respectively, for the four methods.}
  \label{fig:train}
\end{figure}

To analyze key design choices in our differentiable training pipeline, we evaluate four variants at 10,000 training steps (Fig.~\ref{fig:train}): \textbf{LNN-Fly} (ours), \textbf{Diff-Fly}, \textbf{Standard CfC}, and \textbf{LNN-Fly w/o $\lambda$}. During training, the target maximum speed is randomly sampled from $[3, 13)$\,m/s. We report collision-free success rate (safety) and average/maximum speeds (agility). \textbf{LNN-Fly w/o $\lambda$} follows the same setting as \textbf{LNN-Fly} but removes the adaptive forgetting term.

The ablation highlights complementary roles for the two temporal-design choices. Comparing \textbf{LNN-Fly} against \textbf{LNN-Fly w/o $\lambda$} isolates the effect of adaptive forgetting: removing $\lambda$ preserves much of the safety but sharply reduces speed, suggesting that the forgetting gate enables rapid state refresh near hazards while retaining the structured continuous-time core. Comparing \textbf{LNN-Fly w/o $\lambda$} against \textbf{Standard CfC} further supports the structured dynamic-programming-inspired core design. \textbf{Diff-Fly} remains competitive during training, but the transfer results above show that this advantage is less stable under timing and simulator shifts.

\subsection{Real-world Demonstrations}

To validate transferability, the policy trained in the differentiable simulator is deployed to a physical quadrotor without fine-tuning. Sim-to-real consistency is maintained through a unified sector-based LiDAR encoding, where raw Livox Mid-360 point clouds are converted into the sparse sector features used by the policy.

We validate the policy in both outdoor and indoor real-world flights (Fig.~\ref{fig:realfly}). In the outdoor forest, the quadrotor reaches \textbf{4.0\,m/s} while avoiding dense obstacles. In the indoor corridor, it runs at both 10\,Hz and 30\,Hz, maintains stable flight at \textbf{2.0\,m/s}, and achieves \textbf{100\%} success over 20 flights with zero collisions. Table~\ref{tab:realfly_latency} shows that the median perception-and-inference cost stays well below the control periods, while the longer onboard tail mainly reflects occasional real-flight scheduling and communication jitter.

\begin{table}[t]
\caption{\textbf{Latency statistics on desktop and onboard platforms.} LiDAR preprocessing converts raw Livox Mid-360 point clouds into sector features. Desktop measurements are reported on the workstation platform, while onboard measurements are collected during real-flight deployment. We report both the median and the 95th percentile (P95).}
\label{tab:realfly_latency}
\centering
\scriptsize
\setlength{\tabcolsep}{2.5pt}
\renewcommand{\arraystretch}{1.1}
\begin{tabular}{l l c c}
\hline
\textbf{Stage} & \textbf{Platform} & \textbf{Median (ms)} & \textbf{P95 (ms)} \\
\hline
LiDAR preprocessing & desktop workstation & 2.88 & 3.55 \\
LiDAR preprocessing & onboard computer & 3.93 & 8.70 \\
Policy inference & desktop workstation & 0.514 & 0.591 \\
Policy inference & onboard computer (10\,Hz) & 2.48 & 28.31 \\
Policy inference & onboard computer (30\,Hz) & 2.57 & 27.23 \\
\hline
\end{tabular}
\end{table}

\section{CONCLUSIONS, LIMITATIONS AND FUTURE WORK}

We presented \textbf{LNN-Fly}, a deployment-oriented continuous-time UAV navigation framework for LiDAR-based obstacle avoidance in cluttered environments. By combining a \textbf{dynamic-programming-inspired structured recurrent update} with explicit $\Delta t$ conditioning and \textbf{input-driven adaptive forgetting}, LNN-Fly improves obstacle-avoidance tolerance in the tested simulator-transfer, sparse-observation, and control-jitter settings. Trained via \textbf{differentiable physics}, the policy achieves \textbf{sub-millisecond policy-network inference on a desktop GPU}, \textbf{millisecond-level onboard CPU execution}, and zero-shot physical deployment in representative cluttered scenes. These results suggest that task-structured continuous-time policies are a practical direction for reliable learned UAV navigation under deployment-relevant shifts.

\textbf{Limitations.} Despite these gains, performance can still degrade during high-speed flight, where reduced reaction margins amplify residual discretization and actuation delays and ultimately lower the success rate. In addition, the current system does not incorporate an explicit dynamic-obstacle prediction module; consequently, avoidance success drops substantially in scenarios with fast-moving obstacles because the policy has limited ability to anticipate obstacle motion.

\textbf{Future work.} We will further lighten the architecture for resource-constrained onboard deployment and extend the framework to \textbf{dynamic obstacle avoidance} through motion prediction and uncertainty-aware planning.










\bibliographystyle{IEEEtran}
\bibliography{refs}

\begin{thebibliography}{10}
\providecommand{\url}[1]{#1}
\csname url@samestyle\endcsname
\providecommand{\newblock}{\relax}
\providecommand{\bibinfo}[2]{#2}
\providecommand{\BIBentrySTDinterwordspacing}{\spaceskip=0pt\relax}
\providecommand{\BIBentryALTinterwordstretchfactor}{4}
\providecommand{\BIBentryALTinterwordspacing}{\spaceskip=\fontdimen2\font plus
\BIBentryALTinterwordstretchfactor\fontdimen3\font minus
  \fontdimen4\font\relax}
\providecommand{\BIBforeignlanguage}[2]{{%
\expandafter\ifx\csname l@#1\endcsname\relax
\typeout{** WARNING: IEEEtran.bst: No hyphenation pattern has been}%
\typeout{** loaded for the language `#1'. Using the pattern for}%
\typeout{** the default language instead.}%
\else
\language=\csname l@#1\endcsname
\fi
#2}}
\providecommand{\BIBdecl}{\relax}
\BIBdecl

\bibitem{ref25}
E.~Kaufmann, L.~Bauersfeld, A.~Loquercio, M.~M{\"u}ller, V.~Koltun, and
  D.~Scaramuzza, ``Champion-level drone racing using deep reinforcement
  learning,'' \emph{Nature}, vol. 620, no. 7976, pp. 982--987, 2023.

\bibitem{ref9}
Y.~Song, A.~Romero, M.~M{\"u}ller, V.~Koltun, and D.~Scaramuzza, ``Reaching the
  limit in autonomous racing: Optimal control versus reinforcement learning,''
  \emph{Science Robotics}, vol.~8, no.~82, p. eadg1462, 2023.

\bibitem{ref5}
A.~Loquercio, E.~Kaufmann, R.~Ranftl, M.~M{\"u}ller, V.~Koltun, and
  D.~Scaramuzza, ``Learning high-speed flight in the wild,'' \emph{Science
  Robotics}, vol.~6, no.~59, p. eabg5810, 2021.

\bibitem{ref14}
Y.~Zhang, Y.~Hu, Y.~Song, D.~Zou, and W.~Lin, ``Learning vision-based agile
  flight via differentiable physics,'' \emph{Nature Machine Intelligence},
  vol.~7, no.~6, pp. 954--966, 2025.

\bibitem{ref15}
Y.~Hu, Y.~Zhang, Y.~Song, Y.~Deng, F.~Yu, L.~Zhang, W.~Lin, D.~Zou, and W.~Yu,
  ``Seeing through pixel motion: Learning obstacle avoidance from optical flow
  with one camera,'' \emph{IEEE Robotics and Automation Letters}, 2025.

\bibitem{ref1}
C.~Sampedro, A.~Rodriguez-Ramos, H.~Bavle, A.~Carrio, P.~De~la Puente, and
  P.~Campoy, ``A fully-autonomous aerial robot for search and rescue
  applications in indoor environments using learning-based techniques,''
  \emph{Journal of Intelligent \& Robotic Systems}, vol.~95, no.~2, pp.
  601--627, 2019.

\bibitem{ref4}
A.~Kouris and C.-S. Bouganis, ``Learning to fly by myself: A self-supervised
  cnn-based approach for autonomous navigation,'' in \emph{2018 IEEE/RSJ
  International Conference on Intelligent Robots and Systems (IROS)}.\hskip 1em
  plus 0.5em minus 0.4em\relax IEEE, 2018, pp. 1--9.

\bibitem{zhang2024shrinking}
Y.~Zhang, B.~Luo, A.~Mukhopadhyay, D.~Stojcsics, D.~Elenius, A.~Roy, S.~Jha,
  M.~Maroti, X.~Koutsoukos, G.~Karsai \emph{et~al.}, ``Shrinking pomcp: A
  framework for real-time uav search and rescue,'' in \emph{2024 International
  Conference on Assured Autonomy (ICAA)}.\hskip 1em plus 0.5em minus
  0.4em\relax IEEE, 2024, pp. 48--57.

\bibitem{hasan2022uav}
Z.~Hasan, S.~Kumar, V.~Patel, N.~Poplavskyy, P.~Subbaraman, and N.~Xue, ``Uav
  path planning for wildfires-sustainably fighting wildfires with automated
  path planning for uavs,'' 2022.

\bibitem{ref19}
E.~Salvato, G.~Fenu, E.~Medvet, and F.~A. Pellegrino, ``Crossing the reality
  gap: A survey on sim-to-real transferability of robot controllers in
  reinforcement learning,'' \emph{IEEE Access}, vol.~9, pp. 153\,171--153\,187,
  2021.

\bibitem{wu2025model}
H.~Wu, W.~Wang, T.~Wang, and S.~Suzuki, ``Model-free uav navigation in unknown
  complex environments using vision-based reinforcement learning,''
  \emph{Drones}, vol.~9, no.~8, p. 566, 2025.

\bibitem{ref20}
S.~Hochreiter and J.~Schmidhuber, ``Long short-term memory,'' \emph{Neural
  computation}, vol.~9, no.~8, pp. 1735--1780, 1997.

\bibitem{ref35}
A.~Quach, M.~Chahine, A.~Amini, R.~Hasani, and D.~Rus, ``Gaussian splatting to
  real world flight navigation transfer with liquid networks,'' \emph{arXiv
  preprint arXiv:2406.15149}, 2024.

\bibitem{rubanova2019latent}
Y.~Rubanova, R.~T. Chen, and D.~K. Duvenaud, ``Latent ordinary differential
  equations for irregularly-sampled time series,'' \emph{Advances in neural
  information processing systems}, vol.~32, 2019.

\bibitem{ref16}
J.~Pan, J.~Xing, R.~Reiter, Y.~Zhai, E.~Aljalbout, and D.~Scaramuzza,
  ``Learning on the fly: Rapid policy adaptation via differentiable
  simulation,'' \emph{IEEE Robotics and Automation Letters}, 2026.

\bibitem{zhou2019robust}
B.~Zhou, F.~Gao, L.~Wang, C.~Liu, and S.~Shen, ``Robust and efficient quadrotor
  trajectory generation for fast autonomous flight,'' \emph{IEEE Robotics and
  Automation Letters}, vol.~4, no.~4, pp. 3529--3536, 2019.

\bibitem{zhou2020ego}
X.~Zhou, Z.~Wang, H.~Ye, C.~Xu, and F.~Gao, ``Ego-planner: An esdf-free
  gradient-based local planner for quadrotors,'' \emph{IEEE Robotics and
  Automation Letters}, vol.~6, no.~2, pp. 478--485, 2020.

\bibitem{zhou2021ego}
X.~Zhou, J.~Zhu, H.~Zhou, C.~Xu, and F.~Gao, ``Ego-swarm: A fully autonomous
  and decentralized quadrotor swarm system in cluttered environments,'' in
  \emph{2021 IEEE international conference on robotics and automation
  (ICRA)}.\hskip 1em plus 0.5em minus 0.4em\relax IEEE, 2021, pp. 4101--4107.

\bibitem{ref2}
N.~Smolyanskiy, A.~Kamenev, J.~Smith, and S.~Birchfield, ``Toward low-flying
  autonomous mav trail navigation using deep neural networks for environmental
  awareness,'' in \emph{2017 IEEE/RSJ international conference on intelligent
  robots and systems (IROS)}.\hskip 1em plus 0.5em minus 0.4em\relax IEEE,
  2017, pp. 4241--4247.

\bibitem{ref12}
J.~Lu, X.~Zhang, H.~Shen, L.~Xu, and B.~Tian, ``You only plan once: A
  learning-based one-stage planner with guidance learning,'' \emph{IEEE
  Robotics and Automation Letters}, vol.~9, no.~7, pp. 6083--6090, 2024.

\bibitem{ref13}
J.~Lu, Y.~Hui, X.~Zhang, W.~Feng, H.~Shen, Z.~Li, and B.~Tian,
  ``Yopov2-tracker: An end-to-end agile tracking and navigation framework from
  perception to action,'' \emph{arXiv preprint arXiv:2505.06923}, 2025.

\bibitem{ref32}
A.~Bhattacharya, N.~Rao, D.~Parikh, P.~Kunapuli, Y.~Wu, Y.~Tao, N.~Matni, and
  V.~Kumar, ``Vision transformers for end-to-end vision-based quadrotor
  obstacle avoidance,'' in \emph{2025 IEEE International Conference on Robotics
  and Automation (ICRA)}.\hskip 1em plus 0.5em minus 0.4em\relax IEEE, 2025,
  pp. 1--8.

\bibitem{romero2025dream}
A.~Romero, A.~Shenai, I.~Geles, E.~Aljalbout, and D.~Scaramuzza, ``Dream to
  fly: Model-based reinforcement learning for vision-based drone flight,''
  \emph{arXiv preprint arXiv:2501.14377}, 2025.

\bibitem{ref33}
M.~Adang, J.~Low, O.~Shorinwa, and M.~Schwager, ``Singer: An onboard generalist
  vision-language navigation policy for drones,'' \emph{arXiv preprint
  arXiv:2509.18610}, 2025.

\bibitem{ref18}
Z.~Xu, \emph{Reasoning for Representations for Learning-Based Control}.\hskip
  1em plus 0.5em minus 0.4em\relax University of California, Berkeley, 2021.

\bibitem{ref23}
J.~Tobin, R.~Fong, A.~Ray, J.~Schneider, W.~Zaremba, and P.~Abbeel, ``Domain
  randomization for transferring deep neural networks from simulation to the
  real world,'' in \emph{2017 IEEE/RSJ international conference on intelligent
  robots and systems (IROS)}.\hskip 1em plus 0.5em minus 0.4em\relax IEEE,
  2017, pp. 23--30.

\bibitem{ref11}
I.~Akkaya, M.~Andrychowicz, M.~Chociej, M.~Litwin, B.~McGrew, A.~Petron,
  A.~Paino, M.~Plappert, G.~Powell, R.~Ribas \emph{et~al.}, ``Solving rubik's
  cube with a robot hand,'' \emph{arXiv preprint arXiv:1910.07113}, 2019.

\bibitem{ref26}
J.~Xing, L.~Bauersfeld, Y.~Song, C.~Xing, and D.~Scaramuzza, ``Contrastive
  learning for enhancing robust scene transfer in vision-based agile flight,''
  in \emph{2024 IEEE international conference on robotics and automation
  (ICRA)}.\hskip 1em plus 0.5em minus 0.4em\relax IEEE, 2024, pp. 5330--5337.

\bibitem{ref27}
H.~Wang, J.~Xing, N.~Messikommer, and D.~Scaramuzza, ``Environment as policy:
  Learning to race in unseen tracks,'' in \emph{2025 IEEE International
  Conference on Robotics and Automation (ICRA)}.\hskip 1em plus 0.5em minus
  0.4em\relax IEEE, 2025, pp. 11\,333--11\,339.

\bibitem{finn2017model}
C.~Finn, P.~Abbeel, and S.~Levine, ``Model-agnostic meta-learning for fast
  adaptation of deep networks,'' in \emph{International conference on machine
  learning}.\hskip 1em plus 0.5em minus 0.4em\relax PMLR, 2017, pp. 1126--1135.

\bibitem{kumar2021rma}
A.~Kumar, Z.~Fu, D.~Pathak, and J.~Malik, ``Rma: Rapid motor adaptation for
  legged robots,'' \emph{arXiv preprint arXiv:2107.04034}, 2021.

\bibitem{ref28}
L.~Bauersfeld, E.~Kaufmann, P.~Foehn, S.~Sun, and D.~Scaramuzza, ``Neurobem:
  Hybrid aerodynamic quadrotor model,'' \emph{arXiv preprint arXiv:2106.08015},
  2021.

\bibitem{ref29}
M.~Alshiekh, R.~Bloem, R.~Ehlers, B.~K{\"o}nighofer, S.~Niekum, and U.~Topcu,
  ``Safe reinforcement learning via shielding,'' in \emph{Proceedings of the
  AAAI conference on artificial intelligence}, vol.~32, no.~1, 2018.

\bibitem{ref30}
Z.~Xu, X.~Han, H.~Shen, H.~Jin, and K.~Shimada, ``Navrl: Learning safe flight
  in dynamic environments,'' \emph{IEEE Robotics and Automation Letters}, 2025.

\bibitem{ref21}
J.~Ou, X.~Guo, M.~Zhu, and W.~Lou, ``Autonomous quadrotor obstacle avoidance
  based on dueling double deep recurrent q-learning with monocular vision,''
  \emph{Neurocomputing}, vol. 441, pp. 300--310, 2021.

\bibitem{ref36}
R.~T. Chen, Y.~Rubanova, J.~Bettencourt, and D.~K. Duvenaud, ``Neural ordinary
  differential equations,'' \emph{Advances in neural information processing
  systems}, vol.~31, 2018.

\bibitem{kidger2020neural}
P.~Kidger, J.~Morrill, J.~Foster, and T.~Lyons, ``Neural controlled
  differential equations for irregular time series,'' \emph{Advances in neural
  information processing systems}, vol.~33, pp. 6696--6707, 2020.

\bibitem{ref38}
R.~Hasani, M.~Lechner, A.~Amini, D.~Rus, and R.~Grosu, ``Liquid time-constant
  networks,'' in \emph{Proceedings of the AAAI conference on artificial
  intelligence}, vol.~35, no.~9, 2021, pp. 7657--7666.

\bibitem{ref37}
R.~Hasani, M.~Lechner, A.~Amini, L.~Liebenwein, A.~Ray, M.~Tschaikowski,
  G.~Teschl, and D.~Rus, ``Closed-form continuous-time neural networks,''
  \emph{Nature Machine Intelligence}, vol.~4, no.~11, pp. 992--1003, 2022.

\bibitem{ref34}
P.~Kao, ``Robust flight navigation with liquid neural networks,'' Ph.D.
  dissertation, Massachusetts Institute of Technology, 2022.

\bibitem{ref40}
G.~Xu, T.~Wu, Z.~Wang, Q.~Wang, and F.~Gao, ``Flying on point clouds with
  reinforcement learning,'' in \emph{2025 IEEE/RSJ International Conference on
  Intelligent Robots and Systems (IROS)}.\hskip 1em plus 0.5em minus
  0.4em\relax IEEE, 2025, pp. 7231--7238.

\end{thebibliography}

\end{document}